\documentclass[letterpaper, 10 pt, conference]{ieeeconf}  % Comment this line out if you need a4paper

\IEEEoverridecommandlockouts                              % This command is only needed if 
\usepackage{amssymb}
\usepackage{cite}
\usepackage{amsmath,amssymb,amsfonts}
\usepackage{algorithmic}
\usepackage{graphicx}
\usepackage{textcomp}
\usepackage{xcolor}
\usepackage{booktabs} 
\usepackage{multirow} 
\usepackage[caption=false]{subfig}

\usepackage{tabularx}
\usepackage{booktabs}
\usepackage{caption}  % 可选：控制 caption 行距
\usepackage{siunitx}
\usepackage{color}

\title{\LARGE \bf
VLM-Augmented Degradation Modeling for Image Restoration Under Adverse Weather Conditions
}

\author{Qianyi Shao$^{1}$, Yuanfan Zhang$^{2}$, Renxiang Xiao$^{1}$, Liang Hu$^*$$^{1}$
}

\begin{document}

\maketitle
\thispagestyle{empty}
\pagestyle{empty}

%%%%%%%%%%%%%%%%%%%%%%%%%%%%%%%%%%%%%%%%%%%%%%%%%%%%%%%%%%%%%%%%%%%%%%%%%%%%%%%%
\begin{abstract}
Reliable visual perception under adverse weather conditions, such as rain, haze, snow, or a mixture of them, is desirable yet challenging for autonomous driving and outdoor robots. In this paper, we propose a unified Memory‑Enhanced Visual‑Language Recovery (MVLR) model that restores images from different degradation  levels  under various weather conditions. MVLR couples a lightweight encoder–decoder backbone with a Visual‑Language Model (VLM) and an Implicit Memory Bank (IMB). The VLM performs chain‑of‑thought inference to encode weather degradation priors and the IMB stores continuous latent representations of degradation patterns. The VLM-generated priors query the IMB to retrieve fine‑grained degradation prototypes. These prototypes are then adaptively fused with multi-scale visual features via dynamic cross-attention mechanisms, enhancing restoration accuracy while maintaining computational efficiency. Extensive experiments on four severe‑weather benchmarks show that MVLR surpasses single‑branch and Mixture‑of‑Experts baselines in terms of Peak Signal‑to‑Noise Ratio (PSNR) and Structural Similarity Index Measure (SSIM). These results indicate that MVLR offers a practical balance between model compactness and expressiveness for real‑time deployment in diverse outdoor conditions.

\end{abstract}

%%%%%%%%%%%%%%%%%%%%%%%%%%%%%%%%%%%%%%%%%%%%%%%%%%%%%%%%%%%%%%%%%%%%%%%%%%%%%%%%
\section{INTRODUCTION}
Outdoor vision systems for autonomous driving \cite{Hu2023planning} and remote monitoring \cite{Doshi2022Multi} must remain reliable in adverse weather conditions such as rain \cite{qian2018attentive}, fog \cite{He2011single, song2023vision}, and snow \cite{liu2018desnownet}. These atmospheric particles scatter and absorb light, alter low-level image statistics, and degrade visual perception. Classical physics-based priors can reverse a single type of degradation, but fail when multiple weather conditions coexist \cite{chen2023learning, luo2024WMMoE}. Current deep network-based solutions improve versatility across different weather conditions, either using single-branch joint models that compress all degradations into a single parameter set \cite{chen2022learning, li2020all, Radford2021learning}, or multi-branch models that assign a branch network to one kind of weather condition \cite{li2022all, Ozan2023Restoring, Zamir2021}. However, these methods struggle to balance the fine-grained weather classification and the image restoration model size.

The root cause of this dilemma is the lack of explicit degradation modeling under different weather types, such as identifying degradation from images and the degradation level. VLMs can understand and output multidimensional priors of images under different types of degradation models as abstract languages, thereby providing a textual description of the priors of degradation model knowledge \cite{li2022BLIP, li2023BLIP2, Radford2021learning}. However, existing VLM-introducing approaches \cite{languagedriven} typically attempt to discretize degradations into coarse categories and combine them with a mixture of experts (MoE) layer. This strategy only uses VLM as a classifier to guide the image restoration model, and lacks further utilization of the model thinking chain to achieve a detailed understanding of restoration under different scenarios.

We introduce MVLR, which retains a single encoder-decoder backbone while inserting an IMB in the latent space. A global VLM embedding first encodes the spatial localization, degradation type, and severity of the entire image. The embedding then queries the IMB to retrieve implicit degradation prototypes, which capture fine-grained continuous variations without discretization. A cosine similarity module fuses the retrieved prototypes with multi-scale encoder features, and the decoder reconstructs a sharp image.

\begin{figure}
    \centering
    \includegraphics[width=\linewidth]{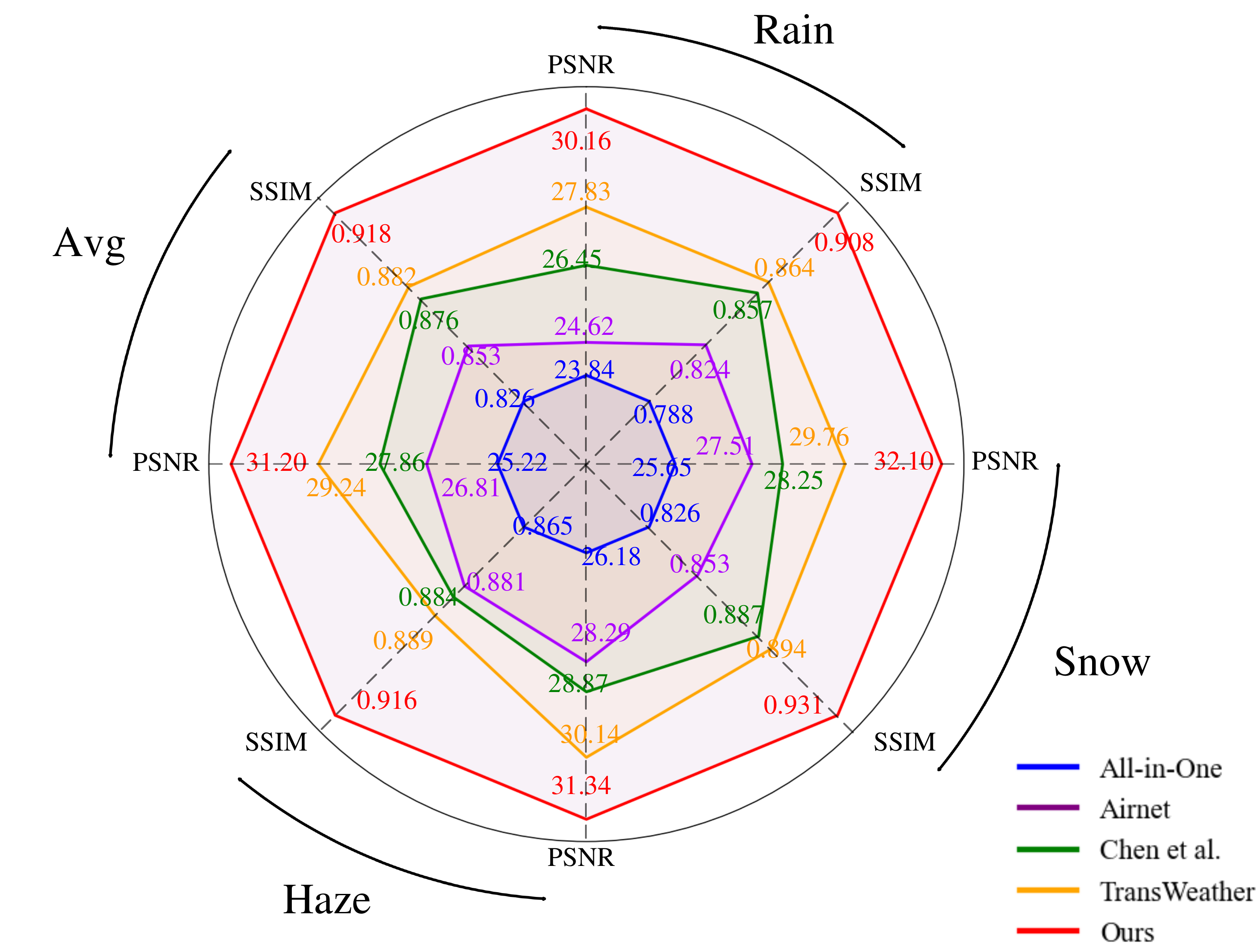}
    \caption{Quantitative comparison (PSNR and SSIM). We present a comparison between our model (red) and baseline methods on three representative degradation scenarios. The superscripts next to the evaluation metrics indicate the corresponding weather degradation type.}
    \label{fig:intro}
\end{figure}

\begin{figure*}
    \centering
    \includegraphics[width=\linewidth]{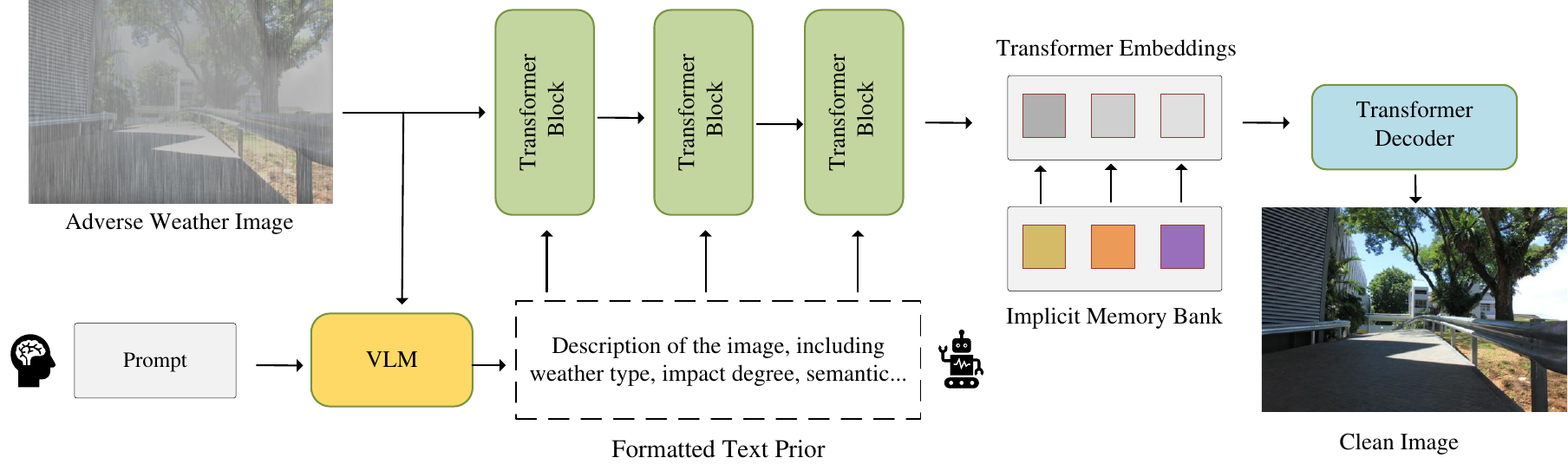}
    \caption{System overview of the MVLR pipeline. Given a degraded image $I^{degraded}$, the model aims to restore a clean image $I^{clean}$. A VLM with prompt words $T^{prompt}$ generates a description embedding $T^{embed}$ that captures weather type, degradation severity, and scene information. This embedding is mapped and fused with image features in the encoder. An implicit memory module then enhances the joint embedding using multi-dimensional degradation prototypes. Finally, the enhanced embeddings are passed through a transformer decoder and a convolution tail to recover the clean image.}
    \label{fig:pipeline}
\end{figure*}

Here are our contributions:
\begin{enumerate}
\item We combine a compact encoder–decoder backbone with a VLM. The VLM emits chain‑of‑thought tokens that localize, classify, and grade weather‑induced degradations, and it steers a prototype‑retrieval module that enriches visual features, yielding high‑fidelity decoding.

\item We propose an IMB that stores degraded prototypes. For each image, semantic labels activate only a few prototypes, enhancing features without the parameter overhead of MoE structures while preserving representational richness.

\item Extensive tests on rain, haze, snow, and mixed‑weather benchmarks show that MVLR surpasses single‑branch and MoE baselines in PSNR and SSIM, confirming its suitability for real‑time adverse‑weather perception pipelines.

\end{enumerate}

As shown in Fig.~\ref{fig:intro}, MVLR achieves higher restoration fidelity than previous joint or multi-scale model architectures, and consistently provides sharper outputs under various weather conditions.

\section{Related Work}

\textbf{Adverse Weather Restoration.}
Early studies on bad weather image restoration trained on a single degradation scene to obtain restoration effects under specific degradation types, such as snow \cite{liu2018desnownet}, rain streaks, raindrops \cite{qian2018attentive} or haze \cite{He2011single}. In order to expand the applicability, methods that use integrated training of multiple weather types have gradually emerged\cite{li2020all, chen2022learning}: some studies use multi-scale CNN (MPRNet \cite{Zamir2021}) or half instance normalization network \cite{chen2021HINet} to better capture rain and snow streaks; others use Transformer (such as Uformer \cite{wang2022Uformer} or Restormer \cite{Zamir2022Restormer}) to model long-range dependencies for image restoration. In addition, Vision Transformer has also been applied to dehazing \cite{song2023vision}. Recent research aims to achieve unified restoration without over-reliance on synthetic data or weather labels\cite{Patil_2023_ICCV, ye2023Adverse, li2022all}. Such methods still implicitly treat each weather type separately, or require additional supervision like known weather categories.

The most recent work \cite{languagedriven} uses VLM for reasoning to enable adaptive restoration using different dynamic subnetworks, however, this approach only focuses on the overall texture consistency covered by particles. Our reasoning process guided by thought chaining can also reason about the relationship between small single objects and surrounding objects even when they are almost completely occluded, and does not degrade performance even under mixed weather degradation or previously unseen weather.

\textbf{Implicit Memory Bank.}
IMB extends the conditional–computation spirit of sparse Mixture‑of‑Experts (MoE) while discarding explicit routing and discrete expert branches. Classic sparse MoE works such as attentive rain routers \cite{qian2018attentive}, probabilistic multilevel MoE framework \cite{Enzweiler2011Multilevel}, large‑scale GLaM \cite{Du2022GLAM}, BotBuster \cite{Ng_Carley_2023}, and weather‑specific Restormer adapters \cite{Zamir2022Restormer} improve capacity by activating only a few specialist networks per input, but they still rely on a gating module and pre‑defined expert identities, causing parameter growth as new conditions are added. Compared with MoE, IMB only needs to activate a few prototypes according to semantic labels by storing diverse degradation prototypes in a shared latent table and retrieved through a step of cosine similarity without the parameter overhead of the MoE structure, while maintaining representation richness.

\textbf{Visual Language Model}
Large pre-trained VLMs, such as BLIP \cite{li2022BLIP}, BLIP-2 \cite{li2023BLIP2}, and CLIP \cite{Radford2021learning}, provide semantically rich image embeddings. Recent studies have injected these embeddings into restoration as auxiliary labels or regularizers\cite{Patil_2023_ICCV, ye2023Adverse}, or used prompts to activate weather experts\cite{Yang_2024_CVPR}. These pipelines only generate text labels for a specific prompt without further reasoning about the details of the image itself. We propose to integrate the mind-chain reasoning results of LLaVA1.5 VLM into IMB queries and collaborate with visual features for enhanced re-driven decoding to achieve adaptive image restoration in different weather conditions.

% \begin{figure*}
%     \centering
%     \includegraphics[width=\linewidth]{Figs/pipeline1.pdf}
%     \caption{System overview of the MVLR pipeline. Given a degraded image $I^{degraded}$, the model aims to restore a clean image $I^{clean}$. A VLM with prompt words $T^{prompt}$ generates a description embedding $T^{embed}$ that captures weather type, degradation severity, and scene information. This embedding is mapped and fused with image features in the encoder. An implicit memory module then enhances the joint embedding using multi-dimensional degradation prototypes. Finally, the enhanced embeddings are passed through a transformer decoder and a convolution tail to recover the clean image.}
%     \label{fig:pipeline}
% \end{figure*}

\section{PROPOSED METHOD}
\subsection{Overview}
The overall framework is shown in Fig.~\ref{fig:pipeline}. Given an image $I^{degraded}$ degraded by bad weather, we aim to restore a clean image $I^{clean}$ through the model. First, our model introduces a visual language model and uses appropriate prompt words $T^{prompt}$ to guide the description $T^{embed}$ of the weather type in the image, the degree of impact, the scene of the image itself, and other information. The description information is projected through a mapping network and fused with image features in the encoder stage. In the stage after the encoder, we designed an implicit memory library to memorize multi-dimensional degradation prototypes and enhance the joint embeddings from the encoder. Finally, the enhanced embeddings are passed through a transformer decoder and a convolution tail to recover the clean image.

\begin{figure}
    \centering
    \includegraphics[width=\linewidth]{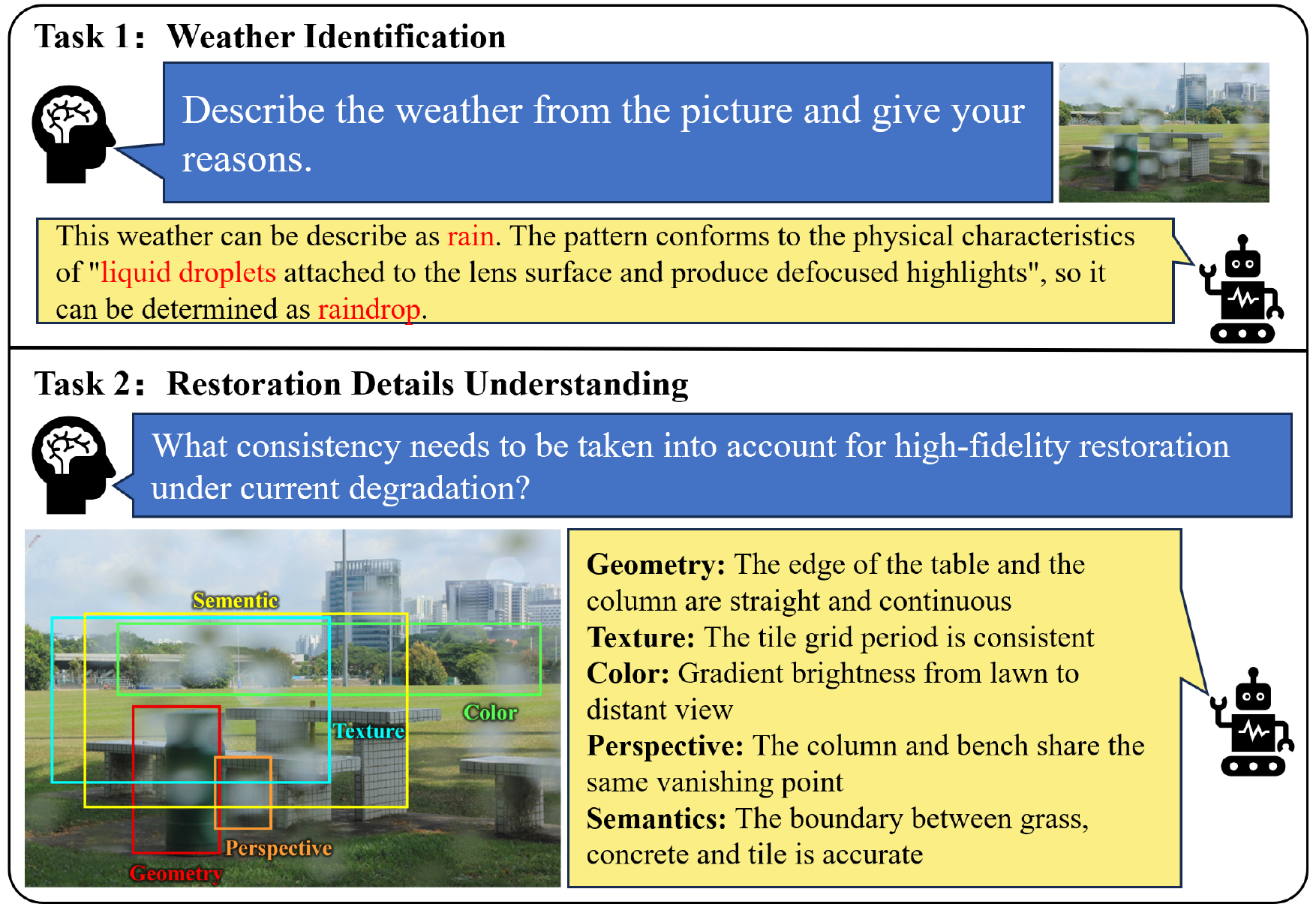}
    \caption{VLM generates a structured text prior process through chain reasoning. The set prompt guides VLM to analyze the degradation type of the environment and further infer about the consistency requirements in the recovery process based on the current degradation situation.}
    \label{fig:VLM}
\end{figure}

\subsection{VLM-Based Degradation Prior}
As shown in Fig.~\ref{fig:VLM}, we utilize the VLM's cross-modal learning ability to infer the prior knowledge of the degraded image. We input the image and the designed prompt into the VLM and obtain the formatted text prior through chain-of-thought reasoning. 
Denote the above process as:
% Through the above process, we can obtain $T^{embed} \in \mathbb{R}^{L \times C^{l} }$:
\begin{equation}
    T^{embed} = VLM(I^{degraded},T^{prompt}), \ T^{embed} \in \mathbb{R}^{L \times C^{l} },
    % T^{embed} = VLM(I^{degraded},T^{prompt}),
\end{equation}
where $VLM(\cdot, \cdot)$ is the visual-language model, $L$ is the length of $T^{embed}$ and $C^{l}$ is the language channel dimension. In order to align with the embedding space of the image features, we project the text prior $T^{embed}$ to the same dimension through a multi-layer perceptron: 
\begin{equation}
    P = MLP(T^{embed}), \ P \in \mathbb{R}^{L \times C^{feat}},
\end{equation}
where $C^{feat}$ is the channel dimension of image features. The projected result forms the final degraded prior $P$ and helps to extract higher-dimensional features in the encoder stage.

\subsection{VLM-driven Transformer Encoder}

We employ a transformer-based encoder to integrate high-level semantic information with local image features, using the degradation prior $P$ from VLM to guide pixel-level restoration of adverse weather conditions by the cross-attention mechanism. The multi-modal fusion is achieved by projecting image features to Query space and degradation prior to Key/Value spaces, followed by attention-based feature aggregation. The effectiveness of fused features is ensured through semantic alignment in attention weights and end-to-end optimization under reconstruction loss supervision. In the specific implementation, we first calculate the semantic alignment between $P$ and image features $F$ and then convert $P$ into a pixel-level degradation representation $X$. The whole process follows the classic $qkv$ attention formula denoted as:
\begin{equation}
    Q=FW^Q,\ K=PW^K,\ V=PW^V,
\end{equation}
\begin{equation}
    X = Attn(Q, K, V) = softmax(\frac{QK^T}{\sqrt{d_k}})V,
\end{equation}
where $W^Q$, $W^K$, $W^V$ are weight matrices for obtaining the Query, Key, and Value, $d_k$ represents the dimensionality of the feature vector $K$.

\subsection{Implicit Memory Bank}
To enhance the model's ability to restore structural details in severely degraded regions, we introduce an implicit memory bank between the encoder and decoder, which retrieves implicit degradation prototypes to guide the reconstruction. The memory bank $M$ is optimized along with the network parameters during training, then frozen and utilized as a static knowledge base during testing. Furthermore, we adopt a Top-$k$ retrieval mechanism that restricts attention computation to the most relevant memory entries to reduce computational cost.

Specifically, we define a learnable memory bank $M = \{m_i\}_{i=1}^{K}$, where each memory slot $m_i \in \mathbb{R}^{C}$ is a trainable vector, randomly initialized and jointly optimized with the network to encode prototypical degradation patterns from diverse training data. This allows the IMB to generalize across types and severity of degradation by retrieving relevant prototypes even for unseen degradations during testing. Unlike external or offline memory systems, our memory is embedded directly into the network as learnable parameters:
\begin{equation}
    M \in \mathbb{R}^{K \times C},
\end{equation}
where $K$ is the number of memory entries and $C$ is the feature dimension.

Given a degradation-aware representation $X \in \mathbb{R}^{H \times W \times C}$, we first extract a global query vector $q \in \mathbb{R}^{C}$ via global average pooling:
\begin{equation}
    q = \text{GAP}(X).
\end{equation}
We then compute the cosine similarity $s_i$ between $q$ and each memory slot $m_i$, and retrieve the top-$k$ most relevant memory entries $M^{\text{top}}$:
\begin{equation}
    s_i = \frac{q^\top m_i}{\|q\| \cdot \|m_i\|}, \quad i = 1, \dots, K,
\end{equation}
\begin{equation}
    M^{\text{top}} = \{ m_{\pi(1)}, m_{\pi(2)}, \dots, m_{\pi(k)} \},
\end{equation}
where $\pi: \{1, \dots, K\} \rightarrow \{1, \dots, K\}$  is a permutation s.t. $ s_{\pi(1)} \ge \cdots \ge s_{\pi(K)}$.

Then we perform a global pooling operation over the selected memory vectors to form a compact prototype vector $m_{proto} \in \mathbb{R}^{C}$, which captures the most relevant external knowledge regarding the current degraded input. Finally, we integrate $m_{proto}$ into the feature map $X$ by broadcasting and residual addition:
\begin{equation}
    \hat{X} = X + \mathbf{1}_{H \times W} \otimes m_{proto}.
\end{equation}
The enhanced feature map $\hat{X}$ contains semantically relevant prototype knowledge retrieved from the learned memory.

\subsection{Loss Function}
To optimize the image restoration network, we adopt a combination of the Charbonnier loss and the Perceptual loss, encouraging both pixel-level accuracy and high-level perceptual quality in the restored images. Given a predicted image $\hat{I}^{clean}$ and its corresponding ground truth $I$, the Charbonnier loss is defined as:
\begin{equation}
    \mathcal{L}_{\text{char}}(I, \hat{I}^{clean}) = \sqrt{\|I - \hat{I}^{clean}\|^2 + \epsilon^2},
\end{equation}
where $\epsilon$ is a small constant of $10^{-3}$ to ensure numerical stability.
We incorporate a perceptual loss based on feature differences extracted from a pre-trained VGG-19 network to further improve the perceptual quality of the restored images. Specifically, we use the activations from intermediate layers of the VGG network to compute the loss:
\begin{equation}
    \mathcal{L}_{\text{perc}}(I, \hat{I}^{clean}) = \sum_{l} \left\| \phi_l(I) - \phi_l(\hat{I}^{clean}) \right\|_2^2,
\end{equation}
where $\phi_l(\cdot)$ denotes the feature map extracted from the $l$-th layer of the VGG network.

The final loss function used to train the network is a weighted sum of the Charbonnier and perceptual losses:
\begin{equation}
    \mathcal{L}_{\text{total}} = \mathcal{L}_{\text{char}} + \lambda \mathcal{L}_{\text{perc}}.
\end{equation}
We set the trade-off parameter $\lambda = 0.05$ to balance pixel fidelity and perceptual quality.

\begin{figure*}
    \centering
    \includegraphics[width=\linewidth]{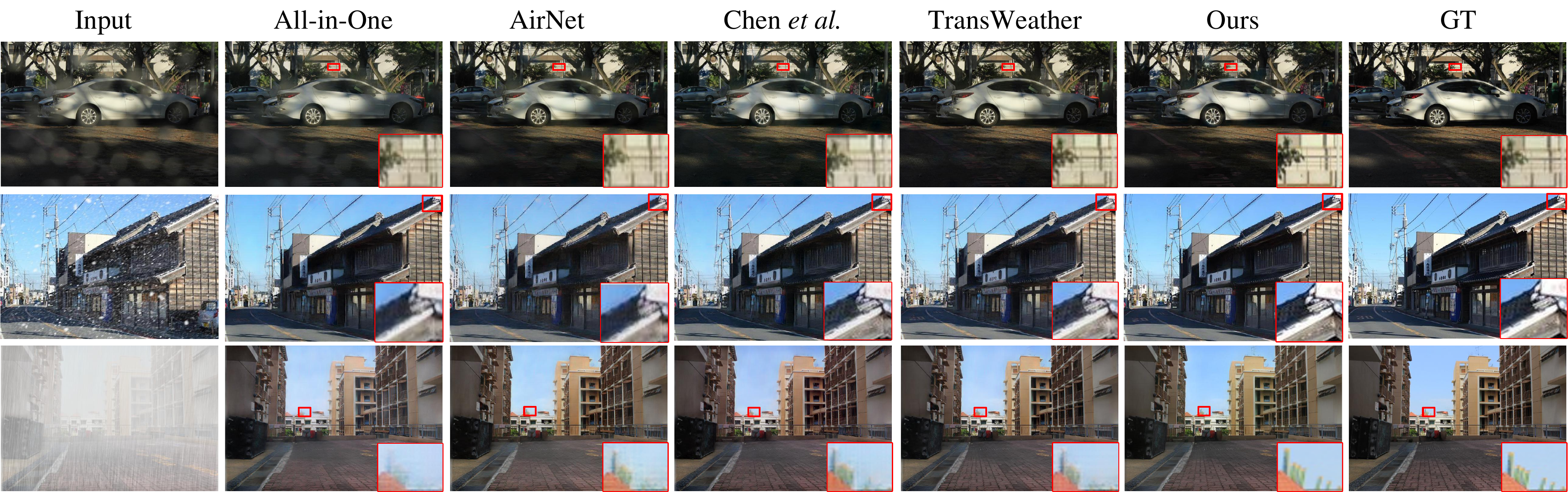}
    \caption{Qualitative comparisons are performed on three representative degradation scenarios (raindrop occlusion, snow streaks, and dense fog). The first column shows the degraded images, and the subsequent columns show the restoration images of state-of-the-art baseline methods (All-in-One, AirNet, Chen \textit{et al.}, and TransWeather) and our method, and the ground truth, with some details magnified.}
    \label{fig:comparasion}
\end{figure*}

% 自定义表头列类型（居中自动换行）
\newcolumntype{Y}{>{\centering\arraybackslash}X}

\begin{table*}[htbp]
\centering
\caption{Visual Quality Comparison between Our Method and Other Adverse Weather Removal Algorithms.}
\label{tab_comp}
\begin{center}
\resizebox{\textwidth}{!}{
%\begin{tabularx}{\textwidth}{lYYYYYYYY}
\begin{tabularx}{\textwidth}{l *{8}{Y} >{\centering\arraybackslash}X}
\toprule
\multirow{2}{*}{Methods} & \multicolumn{2}{c}{\textbf{Rain}} & \multicolumn{2}{c}{\textbf{Snow}} & \multicolumn{2}{c}{\textbf{Haze}} & \multicolumn{2}{c}{\textbf{Average}}\\
\cmidrule(lr){2-3} \cmidrule(lr){4-5} \cmidrule(lr){6-7} \cmidrule(lr){8-9} 
& PSNR & SSIM & PSNR & SSIM & PSNR & SSIM & PSNR & SSIM \\
\midrule
All-in-One      & 23.84 & 0.836 & 25.65 & 0.876 & 26.18 & 0.917 & 25.22 & 0.876\\
Airnet          & 24.62 & 0.873 & 27.51 & 0.904 & 28.29 & 0.934 & 26.81 & 0.904\\
Chen \textit{et al.} & 26.45 & 0.909 & 28.25 & 0.940 & 28.87 & 0.937 & 27.86 & 0.929\\
TransWeather    & \underline{27.83} & \underline{0.916} & \underline{29.76} & \underline{0.948} & \underline{30.14} & \underline{0.942} & \underline{29.24} & \underline{0.935} \\
\textbf{Ours}   & \textbf{30.16} & \textbf{0.963} & \textbf{32.10} & \textbf{0.987} & \textbf{31.34} & \textbf{0.971} & \textbf{31.20} & \textbf{0.973}\\
\bottomrule
\end{tabularx}
}
\end{center}

\vspace{0.5em}
{\raggedright\small Best performance in \textbf{bold}, second best \underline{underlined}.\par}

\end{table*}

\section{EXPERIMENT \& ANALYSIS}
\subsection{Implementation Details}
\paragraph{Experiment Setup}
Our model is implemented using the PyTorch framework and all experiments are conducted on an RTX 4090 GPU. We train our network with a batch size of 8 and $2 \times 10^6$ iterations using the Adam optimizer. The learning rate is decayed from $2 \times 10^{-4}$ to $1 \times 10^{-6}$ by cosine annealing strategy. In training, images are randomly cropped to size $256 \times 256$, and $\lambda = 0.05$. We conduct our experiment on four challenging adverse weather datasets including Outdoor-Rain \cite{li2019heavy}, Raindrop \cite{qian2018attentive}, Snow100K \cite{liu2018desnownet} and RESIDE \cite{li2018benchmarking}.

% 自定义表头列类型（居中自动换行）
\newcolumntype{Y}{>{\centering\arraybackslash}X}

\begin{table*}[htbp]
  \caption{Ablation study on blocks.}
  \label{tab_ablation_block}
  \begin{center}
  \resizebox{\textwidth}{!}{%
  \begin{tabularx}{\textwidth}{lYYYYYYYY}
    \toprule
    \multirow{2}{*}{Methods} & \multicolumn{2}{c}{\textbf{Rain}} & \multicolumn{2}{c}{\textbf{Snow}} & \multicolumn{2}{c}{\textbf{Haze}} & \multicolumn{2}{c}{\textbf{Average}} \\
    \cmidrule(lr){2-3} \cmidrule(lr){4-5} \cmidrule(lr){6-7} \cmidrule(lr){8-9}
     & PSNR & SSIM & PSNR & SSIM & PSNR & SSIM & PSNR & SSIM \\
    \midrule
    Base      & 27.29 & 0.906 & 28.34 & 0.925 & 27.87 & 0.915 & 27.83 & 0.915 \\
    +VLM          & \underline{29.38} & \underline{0.940} & \underline{31.82} & \underline{0.953} & \underline{29.64} & \underline{0.941} & \underline{30.28} & \underline{0.945} \\
    +IMB  & 28.72 & 0.932 & 30.47 & 0.940 & 29.36 & 0.935 & 29.52 & 0.935 \\
    +VLM+IMB(Ours)  & \textbf{30.16} & \textbf{0.963} & \textbf{32.10} & \textbf{0.987} & \textbf{31.34} & \textbf{0.971} & \textbf{31.20} & \textbf{0.973} \\
    \bottomrule
  \end{tabularx}}
  \end{center}
\end{table*}

\paragraph{Baseline Methods and Evaluation Metrics} 
We compare the proposed MVLR with four representative single‑image adverse‑weather removal networks. 
All‑in‑One~\cite{li2020all} is a framework that handles multiple degradations using different encoder parameter sets. 
AirNet\cite{li2022all} employs a dual‑attention encoder–decoder tailored for atmospheric scattering and serves as a strong rain‑to‑haze generalist. Chen \textit{et al.}\cite{chen2022learning} introduce domain‑adaptive feature alignment to cope with different accumulation patterns. 
TransWeather~\cite{valanarasu2022transweather} couples a Transformer backbone with task specific queries and currently represents the state‑of‑the‑art multi‑condition restoration paradigm. 

We adopt PSNR and SSIM as quantitative criteria because they jointly reflect pixel‑level fidelity and perceptual consistency after restoration. 

\subsection{Experimental Results}

\paragraph{Qualitative comparison}
Fig.~\ref{fig:comparasion} compares our method with baseline methods on three representative degradation scenarios. Our proposed method more faithfully restores scene structure, fine-grained texture, and color fidelity than all baseline models.

In the raindrop scene, the output of the baseline method still presents blurred asphalt road and rim outlines, while our method uses VLM to infer the "raindrop + weak specular reflection" prior and retrieves occlusion-aware prototypes from the frozen IMB, thereby separating reflection from illumination and recovering clear brick textures and circular rims.
For the snow scene, the baseline method partially removes snow streaks but leaves unsaturated surfaces. Our network removes all remaining snow streaks and reconstructs high-frequency roof ridges and wooden facade fibers relying on independent snow damage prototypes. In the outdoor-rain scene, VLM classifies this frame as "coexisting rain and mist" and outputs structured prior information such as atmospheric scattering level and local water droplet coverage. In such environments, using separate models for raindrops or mist alone is insufficient for accurate scene interpretation. VLM can accurately distinguish between single weather types and various mixed weather conditions with a 100\% success rate. Guided by these cues, our method successfully achieves multi-scale compensation and near-field occlusion removal, restores distant facade edges, and aligns global illumination and color temperature with ground truth. Experimental synthesis shows that combining the weather-degraded class recognition prior of VLM with the enhanced degradation prototype of IMB can achieve high-quality structure, texture, and color restoration. Even in the case of coupled weather degradation, the best results can still be obtained.

\paragraph{Quantitative Evaluation} 
Tab~\ref{tab_comp} shows that our proposed pipeline attains an average of \SI{31.20}{\decibel} PSNR and \SI{0.973}{} SSIM, exceeding the strongest baseline TransWeather by \SI{1.96}{\decibel} and \SI{0.038}{}, respectively.  The standard deviation of our PSNR gains across the three weather types is \SI{0.54}{\decibel}, demonstrating uniform benefits rather than isolated spikes.  
 
% For Raindrop, our method records \SI{30.16}{\decibel} / \SI{0.908}{} versus TransWeather’s \SI{27.83}{\decibel} / \SI{0.864}{}, a margin of \SI{2.33}{\decibel} and \SI{0.044}{}.  
% Under Moderate Snow, the gains are \SI{2.34}{\decibel} and \SI{0.037}{}.  
% In the mixed challenging Rain and Haze scenario, retrieval from the frozen memory bank supplies an additional \SI{1.20}{\decibel} and \SI{0.027}{}, culminating in \SI{31.34}{\decibel} PSNR and \SI{0.916}{} SSIM.
% \vspace{-0.2em}

% 自定义表头列类型（居中自动换行）
\newcolumntype{Y}{>{\centering\arraybackslash}X}

\begin{table}[htbp]
  \caption{Ablation study on IMB Capacity ($k=\SI{32}{}$).}
  \label{tab_ablation_MB}
  \begin{center}
  
  \begin{tabularx}{.4\textwidth}{lYY}
    \toprule
    Capacity & PSNR & SSIM\\
    \midrule
    64 & 30.32 & 0.952\\
    128 & 30.56 & 0.957\\
    256 & 30.89 & 0.966\\
    512(Ours) & \underline{31.20} & \underline{0.974}\\
    1024 & \textbf{31.27} & \textbf{0.976}\\
    \bottomrule
  \end{tabularx}
  \end{center}

% \vspace{0.5em}
% {\raggedright\small Best performance in \textbf{bold}, second best \underline{underlined}.\par}
\end{table}

% \vspace{-0.5em}
\subsection{Ablation study}
All increments exceed the generally accepted perceptual significance threshold \SI{1}{\decibel}, verifying that the text prior inferred by the visual language model can reliably characterize the degraded physical properties and that feature enhancement after prototype retrieval from the static repository can further refine pixel-level restoration.

\paragraph{Model Architecture}
Tab.~\ref{tab_ablation_block} distinguishes the impact of VLM and IMB. Compared with the original convolutional encoder-decoder baseline, introducing VLM alone can improve PSNR and SSIM to \SI{30.28}{\decibel} and \SI{0.945}.. This shows that visually guided image feature enhancement alone can achieve better decoding results.
The gain of IMB enhancement alone can reach (\SI{29.52}{\decibel}, \SI{0.935}), which confirms that prototype guidance can promote pixel-level recovery even in the absence of explicit semantic cues.
When both cues are used together (Ours), the network achieves \SI{31.20}{\decibel} and \SI{0.973}, which is \SI{3.37}{\decibel} and \SI{0.058} higher than the baseline in terms of PSNR and SSIM, respectively. This synergy suggests that the high-level prior knowledge provided by VLM is complementary to local prototype retrieval, with VLM narrowing the search space to a weather-specific manifold, while IMB provides fine-grained instance-level correction to compensate for local appearance differences.

\paragraph{Memory capacity analysis}
In addition, we fix the search budget to $k=\SI{32}{}$ and vary the capacity $|M| \in {\SI{64}{}, \SI{128}{}, \SI{256}{}, \SI{512}{}, \SI{1024}{}}$ (as to Tab~\ref{tab_ablation_MB}).
Increasing $|M|$ from \SI{64}{} to \SI{512}{}, PSNR steadily improves to \SI{31.20}{\decibel} and SSIM to \SI{0.974}{}.
However, scaling to \SI{1024}{} slots yields only a slight improvement due to diminishing returns once the major weather degradation factors are fully covered.
Therefore, we adopt \SI{512}{} slots as the Pareto-optimal setting: it captures a wide range of archetypes without incurring unnecessary memory or lookup overhead.

\section{CONCLUSIONS}

In this paper, we propose MVLR, a  framework that addresses the  trade-off between compactness and expressiveness in adverse-weather image restoration. By using a global embedding from a VLM, MVLR efficiently captures the spatial distribution, type, and severity of degradations. The embedding then queries the IMB to retrieve implicit degradation prototypes, capturing fine-grained, continuous variations without discretization. A cosine similarity module fuses the retrieved prototypes with multi-scale encoder features, and the decoder reconstructs a clean image. Experiments confirm that MVLR consistently outperforms state-of-the-art baseline methods. 

% \addtolength{\textheight}{-12cm}   % This command serves to balance the column lengths
                                  % on the last page of the document manually. It shortens
                                  % the textheight of the last page by a suitable amount.
                                  % This command does not take effect until the next page
                                  % so it should come on the page before the last. Make
                                  % sure that you do not shorten the textheight too much.
% \addtolength{\textheight}{-3.2cm}   % This command serves to balance the column lengths

%%%%%%%%%%%%%%%%%%%%%%%%%%%%%%%%%%%%%%%%%%%%%%%%%%%%%%%%%%%%%%%%%%%%%%%%%%%%%%%%

%%%%%%%%%%%%%%%%%%%%%%%%%%%%%%%%%%%%%%%%%%%%%%%%%%%%%%%%%%%%%%%%%%%%%%%%%%%%%%%%

%%%%%%%%%%%%%%%%%%%%%%%%%%%%%%%%%%%%%%%%%%%%%%%%%%%%%%%%%%%%%%%%%%%%%%%%%%%%%%%%

\bibliographystyle{ieeetr}  % 你也可以改成 plain、unsrt、abbrv 等
\bibliography{reference}    % 这里不需要加 .bib 后缀

\end{document}